\documentclass[conference]{IEEEtran}
\IEEEoverridecommandlockouts
\usepackage{cite}
\usepackage{amsmath,amssymb,amsfonts,amsthm}
\usepackage{algorithmic}
\usepackage{graphicx}
\usepackage{textcomp}
\usepackage{xcolor}
\usepackage[ruled,vlined,linesnumbered]{algorithm2e}
\usepackage{mathtools}
\usepackage{multirow}
\usepackage{filecontents,lipsum}
\usepackage{subfigure}
\usepackage{tabularx}
\usepackage{array}
\usepackage{url}
\usepackage{threeparttable}

\usepackage[scale=2]{ccicons}
\usepackage{pgfplots}
\usepackage{xspace}
\usepackage{ragged2e}
\usepackage{tabularx}
\usepackage{array}
\usepackage{caption}
\usepackage{xparse}
\usepackage{spot}
\usepackage{mathtools}
\usepackage{bm}

\theoremstyle{definition}
\newtheorem{definition}{Definition}[section]

\def\R{\mathbb{R}}

\def \Xr {X_{\textrm{rref}}}

\def\U{\mathcal{S}_i}
\def\V{\mathcal{S}_j}

\def\dU{d_i}
\def\dV{d_j}

\DeclarePairedDelimiterX\Basics[1](){ #1}

\def\BibTeX{{\rm B\kern-.05em{\sc i\kern-.025em b}\kern-.08em
    T\kern-.1667em\lower.7ex\hbox{E}\kern-.125emX}}
\begin{document}

\title{Subspace Clustering of Very Sparse High-Dimensional Data}

\author{\IEEEauthorblockN{Hankui Peng}
	\IEEEauthorblockA{\textit{STOR-i Centre for Doctoral Training} \\
		\textit{Lancaster University}\\
		Lancaster, UK \\
		h.peng3@lancaster.ac.uk}\\

	\IEEEauthorblockN{Idris Eckley}
	\IEEEauthorblockA{\textit{Department of Mathematics and Statistics} \\
		\textit{Lancaster University}\\
		Lancaster, UK \\
		i.eckley@lancaster.ac.uk}\\
	\and
	\IEEEauthorblockN{Nicos Pavlidis}
	\IEEEauthorblockA{\textit{Department of Management Science} \\
		\textit{Lancaster University}\\
		Lancaster, UK \\
		n.pavlidis@lancaster.ac.uk}\\
	
	\IEEEauthorblockN{Ioannis Tsalamanis}
	\IEEEauthorblockA{\textit{Data Science Campus} \\
		\textit{Office for National Statistics}\\
		Newport, UK \\
		ioannis.tsalamanis@ons.gov.uk}
}

\maketitle

\begin{abstract}
In this paper we consider the problem of clustering collections of very short
texts using subspace clustering. This problem arises in many applications
such as product categorisation, fraud detection, and sentiment analysis.  The main challenge lies in the fact that the vectorial representation of short texts is both high-dimensional, due to the large number of unique terms in the
corpus, and extremely sparse, as each text contains a very small number of
words with no repetition. We propose a new, simple subspace
clustering algorithm that relies on linear algebra to cluster such datasets.
Experimental results on identifying product categories from product names
obtained from the US Amazon website indicate that the algorithm can be
competitive against state-of-the-art clustering algorithms.

\end{abstract}


\section{Introduction}


In recent years, there has been an increasing need to understand and analyse the huge
volumes of text data that have become available on different platforms. For
example, Amazon may wish to automate their product categorisation based on
product names and descriptions, Twitter may wish to utilise automatic online
policing to identify sensitive and non-sensitive Tweets, etc. 
%
%
The lack of labels for the vast majority of such texts makes this an
unsupervised learning problem.

In this work, we study the problem of clustering collections of very short texts.
Short length has two important implications. First, in each
``document'' each word is effectively observed once. Second, the vast majority
of pairs of texts have no words in common.
These properties pose challenges for established text mining algorithms, as
well as for statistical methods that employ generative models, such as the
Latent Dirichlet Allocation~\cite{blei2003latent}, which require long texts to
achieve reliable parameter estimates.

If one uses the standard Term Frequency--Inverse Document
Frequency~\cite{larson2010introduction} (TF-IDF) representation, such
document collections give rise to high-dimensional and very sparse datasets.
In this setting it is sensible to argue that texts sharing even a few number
of common words are very similar to each other. Therefore associating
clusters with linear combinations of the features (i.e. linear subspaces) is reasonable. 

Subspace clustering refers to a set of methods that aim to identify clusters
defined in linear and / or affine subspaces of the full-dimensional data. Such methods can be categorised into four classes: algebraic, statistical, iterative and spectral. {\em Algebraic} methods rely on matrix factorisation~\cite{gear1998multibody}
or polynomial algebra~\cite{vidal2003generalized} to identify subspaces. A
highly cited method from this class is the robust subspace segmentation by low
rank representation~\cite{liu2010robust} (LRR). LRR
%
%
relies on the idea that observations from the same cluster can be expressed by
the same set of bases vectors, and thus as linear combinations of
each other. It first builds a similarity matrix for the data through solving an
optimisation problem based on the low rank data representation, and then
applies spectral clustering.

{\em Statistical} methods impose explicit assumptions about the data generating process
for each cluster. A probabilistic model is estimated based on the principles
of maximum likelihood. A prominent method from this class is the mixture
of probabilistic principal component analysers~\cite{tipping1999mixtures},
which models each cluster as a multivariate Gaussian distribution.

%

{\em Iterative} methods refine the cluster assignment and the estimated
subspaces to optimise an objective function.
Projective $k$-means~\cite{agarwal2004k} (PKM) is such an extension
of the classic $k$-means algorithm. PKM aims to minimise the
root mean square error between each observation and its projection onto
the corresponding low-dimensional subspace. This is achieved by alternating between computing the
cluster centroids and updating the cluster assignment. In PKM the centroid of each cluster is
the mean of the data projected onto a linear subspace defined by the principal
components vectors.


{\em Spectral (clustering)--based} methods construct a similarity matrix that is
representative of how close each pair of data objects are, and then apply standard
spectral clustering. The success of such methods critically depends
on the choice of the similarity measure. 
Sparse Subspace Clustering~\cite{elhamifar2013sparse} (SSC) estimates the similarity matrix by
solving an optimisation problem which aims to express
each observation as a linear combination of the
other observations. The coefficients of the optimal combination
are used in the similarity matrix.

In this paper we propose a new, simple subspace clustering algorithm, 
motivated by the characteristics of short texts. The algorithm first identifies
subspaces that contain few but very similar observations. 
Then an appropriate dissimilarity measure is used to merge these
subspaces into meaningful clusters.
We apply the algorithm on a dataset of product names obtained from Amazon website and made available by the 
The Billion Prices Project~\cite{DVN/XXOUHF_2016},
and show that its performance is competitive with state-of-the-art (subspace) clustering
algorithms.

The rest of the paper is organised as follows. Section~\ref{pf} presents the methodology,
and the comparative evaluation of the proposed algorithm is provided in Section~\ref{experiments}.
Conclusions and future research directions are discussed in Section~\ref{conclusions}.

\section{Methodology}\label{pf}
We obtain vectorial representation for the $N$ product names through the well established Term Frequency--Inverse Document Frequency (TF-IDF)~\cite{larson2010introduction} approach. 
Since each text is very short, and different texts contain different words, the TF-IDF matrix, $X \in \R^{P \times N}$, is sparse and high-dimensional.
%
%
For the specific dataset we consider, each product name consists of a very
small number of words (with effectively no repetition), and the vast majority of pairs of
product names have no words in common.

In subspace clustering, each observation is assumed to lie on (or close to) a
relatively low-dimensional subspace.
A $d_k$-dimensional linear subspace, $S_k \subset \R^P$ is defined as,
\[\mathcal{S}_{k} = \left\{ x \in \R^P \,:\, x= U_{k} y \right\},\]
where $U \in \R^{P \times d_k}$ is an orthonormal matrix defining the basis of
the subspace, 
%
%
and $y \in \R^{d_k}$ is the representation of $x$ in terms of the
column vectors of $U_k$.
%
%
%
%
%
The goal of subspace clustering is to identify the~$K$ subspaces, and assign
each observation to the appropriate subspace. 
In the context of our problem,
features of $X$
%
%
correspond to unique words. It is therefore
sensible to assume that texts that share a
combination of words are similar to each other.

The first step in the proposed approach is to transform the TF-IDF matrix,~$X$,
into its reduced row echelon form~\cite{GolubL1996},
by applying the well
known Gauss-Jordan elimination. In this process a sequence of
row operations are performed to bring~$X$ into a form
that satisfies\footnote{see~\cite{GolubL1996} for numerically stable algorithms to perform this operation}:

\begin{enumerate}

\item the leftmost non-zero entry of each row is~1;

\item the leftmost non-zero entry of each row is the only non-zero entry in the corresponding column;

\item for any two different leftmost non-zero entries, one located in row $i$, column $j$ and the other in row $s$, column $t$. If $s>i$, then $t>j$;

\item rows in which every entry is zero are beneath all rows with non-zero entries. 

\end{enumerate}
Let $\Xr$ denote the reduced row echelon form of~$X$. The columns of $\Xr$ that
have a single non-zero element are called {\em pivot columns}\/. The first column of~$\Xr$ is always a
pivot column. Moreover, column $j>1$ is a pivot column, if and only if 
the $j$-th column of $X$ cannot be expressed as a linear combination of
the previous columns (i.e. columns $1,\ldots,j-1$).
%
%
%
Let $j$ be a non-pivot column of $\Xr$. The non-zero elements in this
column specify the coefficients of the linear combination of the
previous pivot columns that yield the $j$-th column vector.

Since observations that can be written as linear combinations of each other
belong to the same linear subspace, $\Xr$
provides valuable information to identify clusters spanning different
subspaces~\cite{gruber2004multibody}.
A simple approach to identify subsets of observations that belong
to the same linear subspace through $\Xr$ is the following. Define the matrix $Y \in
\{0,1\}^{P\times N}$ as $Y(i,j) = \mathbf{1} \left(\Xr(i,j) \neq 0 \right)$,
where $\mathbf{1}(\cdot)$ is the indicator function that returns one if its
argument is true and zero otherwise.
Then the adjacency matrix, $A = Y^\top Y$, defines a graph, $G(A)$, whose
connected components are subsets of observations that can be expressed as linear combinations of each other.

\begin{figure}[h!]
\centering
\includegraphics[width=.5\textwidth]{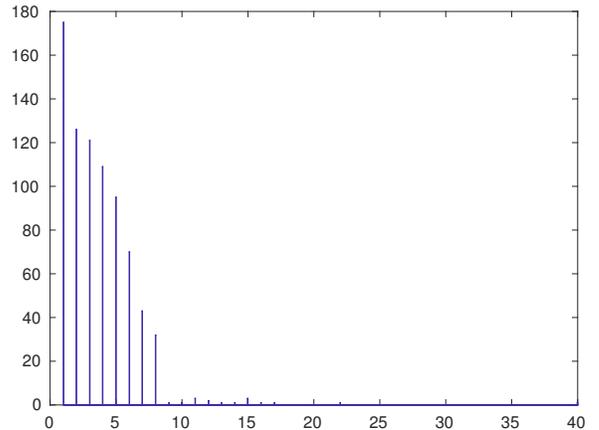}
\caption{Histogram of the number of observations in each subspace identified
through the reduced row echelon form of the TF-IDF matrix.}
\label{fig1}
\end{figure}

For the problem of clustering very short texts the graph
$G(A)$ has a very 
large number of connected components, many of which consist of a single observation.
Texts that belong to the same connected component are very similar, and hence this
partitioning is very accurate in terms of purity~\cite{ZhaoK2004}, but it is of no
practical use since it completely fails to capture broader groups.
Figure~\ref{fig1} provides the histogram of the number of observations in
each connected component of $G(A)$ for the Amazon product names dataset.
As the figure shows, the vast majority of connected components contain less than
ten observations, while the mode of this distribution is at one. 

To form meaningful clusters in such datasets, we need an 
an appropriate measure of dissimilarity that would allow us to merge
the previously identified subspaces.
In this work, we utilise the concept of
\emph{principal angles}, first introduced in~\cite{jordan1875essai}.
%
%

\begin{definition}[Principal Angles]

Let $\U$ and $\V$ be two linear subspaces of an inner product space
with $1 \leqslant \text{dim}\,\U= \dU \leqslant \text{dim}\,\V =
\dV$.
The {\em principal angles}, 
\[ 0\leqslant \theta_{1} \leqslant \theta_{2}\leqslant \ldots \leqslant \theta_{\dU} \leqslant \pi/2, \] 
between $\U$ and $\V$ can be defined recursively for $k = 1,\ldots,\dU$ as,
\[
\cos(\theta_k) = \max_{u \in \U} \max_{v \in \V} \cos( u^\top v) = u_k^\top v_k, \text{ } 
\]
subject to,
\[
\|u\| = \|v\| = 1, \; \textrm{and} \; u^\top u_m = 0, \; v^\top v_m=0, \; \textrm{for} \; 0<m <k.
\]
\end{definition}

%
Applying Principal Component Analysis (PCA) to the subset of observations assigned to each
connected component of $G(A)$, one readily obtains an orthonormal basis for each subspace.
Let the columns of matrices $Q_{\U} \in \R^{P \times \dU}$ and $Q_{\V} \in
\R^{P \times \dV}$ constitute orthonormal bases for two linear subspaces $\U$ and $\V$, respectively.
The principal angles between $\U$ and $\V$
can be obtained from the singular value decomposition, 
$Q_{\U}^\top Q_{\V} = Y \Sigma Z^\top$, as follows
\begin{equation}
\theta_k = \arccos( \Sigma(k,k) ), \; i\in \left\{1,\ldots,\dU\right\}.
\end{equation}
%
%
%
Principal angles ignore the difference in dimensionality between the
two subspaces, which for our purposes is very important.
To accommodate for this, we assume that $\U$ and $\V$ have maximum dissimilarity
along the dimensions $(\dV - \dU)$.
Thus we define the dissimilarity between two linear subspaces, $\U$ and
$\V$ as,
\begin{align}\label{eq:dis}
\text{D}(i,j) &= 
\frac{1}{\dV}\left(\dV- \dU + \sum_{i=1}^{\dU}(1 -\cos( \theta_{i})) \right), \nonumber\\ 
& = 1 - \frac{1}{\dV}\sum_{i=1}^{\dU}\cos( \theta_{i}).
\end{align}

To obtain the final set of~$K$ clusters we apply
the spectral clustering algorithm of Ng et al.~\cite{NgJW2002} using $D$
as the dissimilarity matrix.
This spectral clustering algorithm uses the Gaussian kernel on
pairwise distances / dissimilarities, as such its performance depends on the choice
of the bandwidth parameter. In this work we use the local scaling rule proposed
in~\cite{Zelnik2004},
\begin{equation}\label{eq:W}
W(i,j) = \exp \left\{ - \frac{D(i,j)^2}{ s_i s_j} \right\}, 
\end{equation}
where $s_i$ ($s_j$) is the dissimalirity of the $i$-th ($j$-th) observation to its $k$-th nearest
neighbour.
All the observations allocated to a given subspace are assigned to the same cluster label
as the subspace.
Algorithm~\ref{alg1} outlines the steps of the proposed approach.

\begin{algorithm}
\caption{Minimum Angle Clustering (MAC)}\label{alg1}
\DontPrintSemicolon
\SetAlgoLined
\SetKwInOut{Input}{Input}
\SetKwInOut{Output}{Output}
\Input{TF-IDF matrix $X \in \R^{P \times N}$; Number of clusters~$K$}
\Output{Cluster assignment $\mathcal{C} \in\left\{1,\ldots,K \right\}$}
\BlankLine

Compute Reduced Row Echelon Form: $\Xr = \mathrm{rref}(X)$\;
Define matrix $Y$ through $Y(i,j) = \mathbf{1}(\Xr(i,j) \neq 0)$\;
Construct graph: $G$ from adjacency matrix $A=Y^\top Y$\;
Compute connected components of $G$: $\{c_1,\ldots,c_{n_c}\}$ \;
\For{$i = 1$ \KwTo$n_c$}{
	Apply PCA to $X(:,c_i)$ to obtain orthonormal basis for $i$-th subspace $Q_i \in \R^{P\times d_i}$\;
	\For{$j=1$ \KwTo $i-1$}{
		Estimate dissimilarity with previous subspaces, $D(i,j)$ through Eq.~(\ref{eq:dis})
	}
}

Apply Spectral Clustering on $W$ defined in Eq.~(\ref{eq:W}) to obtain cluster assignment
of subspaces\;

To all the observations in each connected component of $G(A)$ assign the same cluster label as that
of the associated subspace\;

\end{algorithm}

\section{Experimental Results} \label{experiments}
\begin{table*}
	\begin{center}
		\begin{tabular}{|l|r|r|r|r|r|r|r|r|}
			\hline
			{\bf Method} & {\bf MAC}    & {\bf SSC} & {\bf LRR} & {\bf PKM} & {\bf SC($X$)} & {\bf SC($A$)} & {\bf LDA} & {\bf PDDP} \\
			\hline
			Purity       & {\bf 0.742} & 0.219    & 0.510    & 0.591    & 0.512        & 0.519        & 0.510    & 0.578\\
			NMI          & {\bf 0.328} & 0.032    & 0.041    & 0.218    & 0.022        & 0.052        & 0.021    & 0.084\\
			ARI          & {\bf 0.251} & 0.025    & -0.023   & 0.191    & 0.000        & -0.024       & 0.011    & 0.065\\
			\hline
			Runtime &137.672&421.231&3050.412&148.141&6.652&96.688&\textbf{4.624}&15.713\\
			\hline
		\end{tabular}
	\end{center}
	\caption{Clustering performance and runtime comparison (in seconds) on US Amazon web-scraped dataset.}
	\label{results}
\end{table*}

In this section, we compare the performance of our proposed method against state-of-the-art subspace, and standard clustering algorithms on the task of clustering Amazon product names dataset \cite{DVN/XXOUHF_2016}. This dataset contains five broad product categories: Electronics,
Home and appliances, Mix, Office products, and Pharmacy and Health. We use the standard TF-IDF format to represent the product names. The resulting TF-IDF matrix has 2921 observations and  and 2106 features/unique words.
%


%
%

%

We compare the performance of MAC with the following
clustering algorithms: Sparse Subspace Clustering~\cite{elhamifar2013sparse}
(SCC), Low Rank Representation~\cite{liu2010robust} (LRR), Projective $k$-Means
Clustering \cite{agarwal2004k} (PKM),
Spectral Clustering~\cite{NgJW2002} (SC), and Principal Component Divisive
Partitioning~\cite{boley1998principal} (PDDP). 
SSC, LRR and PKM, are state-of-the-art subspace clustering algorithms. PDDP is included
as it has been developed for
the purpose of partitioning documents that have been embedded in high-dimensional
Euclidean space. SC is a generic clustering methodology that has been
successfully applied on numerous high-dimensional applications, most notably
image segmentation~\cite{shi2000normalized}. 
A further reason for including this algorithm in the performance comparison
is that MAC employs SC at its last step.
Thus we need to investigate first, whether the performance of our algorithm is attributable to SC; and 
second whether the information from the connected components of $G(A)$ suffices to correctly identify the clusters 
in this dataset (and hence the
next step of defining dissimilarity based on principal angles is not worthwhile).
%
For completeness we also
consider Latent Dirichlet Allocation~\cite{blei2003latent} (LDA) which has been
widely applied in text mining.

We assess performance through three external cluster evaluation measures:
Purity~\cite{ZhaoK2004}, Normalised Mutual Information~\cite{StrehlG2002}
(NMI), and Adjusted Rand Index~\cite{hubert1985comparing} (ARI).
For all three measures higher values indicate superior performance in the sense
that cluster labels are in better agreement with the actual cluster assignment.
Purity and NMI assume values in $[0,1]$, while the adjusted Rand index is in
$[-1,1]$.
Table~\ref{results} reports the performance of all algorithms on our dataset.
As the table shows, MAC outperforms the other methods with respect
to all three measures. It is important to note that the performance of MAC is
substantially better than that of the two SC variants, the first using the original
TF-IDF data representation (column SC($X$)
in the table), and the second using 
as similarity matrix
the adjacency matrix $A$ obtained after transforming the matrix into the reduced row echelon form.
The second best performing method is PKM, while the purity scores for
PDDP and the two SC variants are comparable to that of PKM.
With the exception of PKM, MAC achieves an improvement of an order of magnitude
compared to all other algorithms with respect to NMI and ARI.
%

\section{Conclusions}\label{conclusions}  

We proposed a new, simple algorithm for subspace clustering that is effective
in clustering collections of very short texts. The algorithm is designed to
exploit the properties of the very sparse and high-dimensional TF-IDF
representation of such datasets. It first identifies low-dimensional linear
subspaces that contain small clusters of texts that share common words. To
merge these into meaningful clusters we use principal angles to quantify the
dissimilarity between linear subspaces, which in the present context correspond
to combinations of words.
Experimental results on a dataset of product names show that this simple approach compares
favourably with standard and subspace clustering methods.

In future work, we aim to develop approaches to correctly identify the hierarchical structure of product categories. 
We also aim to investigate active learning approaches to assist the cluster validation process.

\section*{Acknowledgements}

This research has made use of The Billion Prices Project Dataverse, which is
maintained by Harvard University \& MIT. 
Hankui Peng acknowledges the financial support of Lancaster University and the Office for National Statistics Data Science Campus as part of the EPSRC-funded STOR-i Centre for Doctoral Training.

\bibliographystyle{IEEEtran}

\begin{thebibliography}{10}
	\providecommand{\url}[1]{#1}
	\csname url@samestyle\endcsname
	\providecommand{\newblock}{\relax}
	\providecommand{\bibinfo}[2]{#2}
	\providecommand{\BIBentrySTDinterwordspacing}{\spaceskip=0pt\relax}
	\providecommand{\BIBentryALTinterwordstretchfactor}{4}
	\providecommand{\BIBentryALTinterwordspacing}{\spaceskip=\fontdimen2\font plus
		\BIBentryALTinterwordstretchfactor\fontdimen3\font minus
		\fontdimen4\font\relax}
	\providecommand{\BIBforeignlanguage}[2]{{%
			\expandafter\ifx\csname l@#1\endcsname\relax
			\typeout{** WARNING: IEEEtran.bst: No hyphenation pattern has been}%
			\typeout{** loaded for the language `#1'. Using the pattern for}%
			\typeout{** the default language instead.}%
			\else
			\language=\csname l@#1\endcsname
			\fi
			#2}}
	\providecommand{\BIBdecl}{\relax}
	\BIBdecl
	
	\bibitem{blei2003latent}
	D.~M. Blei, A.~Y. Ng, and M.~I. Jordan, ``Latent dirichlet allocation,''
	\emph{Journal of machine Learning research}, vol.~3, no. Jan, pp. 993--1022,
	2003.
	
	\bibitem{larson2010introduction}
	R.~R. Larson, ``Introduction to information retrieval,'' \emph{Journal of the
		American Society for Information Science and Technology}, vol.~61, no.~4, pp.
	852--853, 2010.
	
	\bibitem{gear1998multibody}
	C.~W. Gear, ``Multibody grouping from motion images,'' \emph{International
		Journal of Computer Vision}, vol.~29, no.~2, pp. 133--150, 1998.
	
	\bibitem{vidal2003generalized}
	R.~Vidal, Y.~Ma, and S.~Sastry, ``Generalized principal component analysis
	(gpca),'' in \emph{2003 IEEE Computer Society Conference on Computer Vision
		and Pattern Recognition, 2003. Proceedings.}, vol.~1.\hskip 1em plus 0.5em
	minus 0.4em\relax IEEE, 2003, pp. I--I.
	
	\bibitem{liu2010robust}
	G.~Liu, Z.~Lin, and Y.~Yu, ``Robust subspace segmentation by low-rank
	representation,'' in \emph{Proceedings of the 27th international conference
		on machine learning (ICML-10)}, 2010, pp. 663--670.
	
	\bibitem{tipping1999mixtures}
	M.~E. Tipping and C.~M. Bishop, ``Mixtures of probabilistic principal component
	analyzers,'' \emph{Neural computation}, vol.~11, no.~2, pp. 443--482, 1999.
	
	\bibitem{agarwal2004k}
	P.~K. Agarwal and N.~H. Mustafa, ``$k$-means projective clustering,'' in
	\emph{Proceedings of the twenty-third ACM SIGMOD-SIGACT-SIGART symposium on
		Principles of database systems}.\hskip 1em plus 0.5em minus 0.4em\relax ACM,
	2004, pp. 155--165.
	
	\bibitem{elhamifar2013sparse}
	E.~Elhamifar and R.~Vidal, ``Sparse subspace clustering: Algorithm, theory, and
	applications,'' \emph{IEEE transactions on pattern analysis and machine
		intelligence}, vol.~35, no.~11, pp. 2765--2781, 2013.
	
	\bibitem{DVN/XXOUHF_2016}
	A.~Cavallo, ``Cavallo (2017) "are online and offline prices similar? evidence
	from large multi-channel retailers" - american economic review - vol. 107(1),
	p.283–303,'' 2016.
	
	\bibitem{GolubL1996}
	G.~H. Golub and C.~F. van Loan, \emph{Matrix Computations (3rd Ed.) [ripped by
		sabbanji]}, 3rd~ed.\hskip 1em plus 0.5em minus 0.4em\relax The John Hopkins
	University Press, 1996.
	
	\bibitem{gruber2004multibody}
	A.~Gruber and Y.~Weiss, ``Multibody factorization with uncertainty and missing
	data using the em algorithm,'' in \emph{Computer Vision and Pattern
		Recognition, 2004. CVPR 2004. Proceedings of the 2004 IEEE Computer Society
		Conference on}, vol.~1.\hskip 1em plus 0.5em minus 0.4em\relax IEEE, 2004,
	pp. I--I.
	
	\bibitem{ZhaoK2004}
	Y.~Zhao and G.~Karypis, ``Empirical and theoretical comparisons of selected
	criterion functions for document clustering,'' \emph{Machine Learning},
	vol.~55, no.~3, pp. 311--331, 2004.
	
	\bibitem{jordan1875essai}
	C.~Jordan, ``Essai sur la g{\'e}om{\'e}triean dimensions,'' \emph{Bull. Soc.
		Math. France}, vol.~3, pp. 103--174, 1875.
	
	\bibitem{NgJW2002}
	A.~Ng, M.~I. Jordan, and Y.~Weiss, ``On spectral clustering: analysis and an
	algorithm,'' in \emph{Advances in Neural Information Processing Systems 14},
	T.~Dietterich, S.~Becker, and Z.~Ghahramani, Eds.\hskip 1em plus 0.5em minus
	0.4em\relax MIT Press, Cambridge, 2002, pp. 849 --856.
	
	\bibitem{Zelnik2004}
	L.~Zelnik-Manor and P.~Perona, ``Self-tuning spectral clustering,'' in
	\emph{Advances in neural information processing systems}, 2004, pp.
	1601--1608.
	
	\bibitem{boley1998principal}
	D.~Boley, ``Principal direction divisive partitioning,'' \emph{Data mining and
		knowledge discovery}, vol.~2, no.~4, pp. 325--344, 1998.
	
	\bibitem{shi2000normalized}
	J.~Shi and J.~Malik, ``Normalized cuts and image segmentation,'' \emph{IEEE
		Transactions on pattern analysis and machine intelligence}, vol.~22, no.~8,
	pp. 888--905, 2000.
	
	\bibitem{StrehlG2002}
	A.~Strehl and J.~Ghosh, ``Cluster ensembles---a knowledge reuse framework for
	combining multiple partitions,'' \emph{Journal of Machine Learning Research},
	vol.~3, pp. 583--617, 2002.
	
	\bibitem{hubert1985comparing}
	L.~Hubert and P.~Arabie, ``Comparing partitions,'' \emph{Journal of
		classification}, vol.~2, no.~1, pp. 193--218, 1985.
	
\end{thebibliography}

\end{document}